\documentclass[conference,9pt]{IEEEtran}
\IEEEoverridecommandlockouts
\usepackage{adjustbox}
\usepackage{cite}
\usepackage{amsmath,amssymb,amsfonts}
\usepackage{algorithmic}
\usepackage{graphicx,subcaption}
\usepackage{textcomp}
\usepackage{multirow}
\usepackage{xcolor}
\usepackage{tikz}
\usepackage{pifont}
\usepackage{booktabs}
\usepackage{stfloats} 
\usepackage{array}
\usepackage[a4paper, total={184mm,239mm}]{geometry}
\usepackage{comment}
\usepackage{url}
\def\BibTeX{{\rm B\kern-.05em{\sc i\kern-.025em b}\kern-.08em
    T\kern-.1667em\lower.7ex\hbox{E}\kern-.125emX}}
\usepackage{array}
\newcolumntype{P}[1]{>{\centering\arraybackslash}p{#1}} 
 \usepackage{xspace}
\newcommand{\pp}{\,pp\xspace}
\usepackage{makecell}

\definecolor{yellowstar}{HTML}{FFA503}

\usepackage{fancyhdr}

\fancypagestyle{firstpage}{%
\fancyhf{}

 \lhead{This work has been accepted at the IEEE Computer Society Annual Symposium on VLSI (ISVLSI) 2026}
 }

\begin{document}

\bstctlcite{BSTcontrol}

\title{AxMoE: Characterizing the Impact of Approximate Multipliers on Mixture-of-Experts DNN Architectures } 

\author{
\centering
\begin{tabular}{ccc}
\textbf{Omkar B Shende} & \textbf{Marcello Traiola} & \textbf{Gayathri Ananthanarayanan} \\

\textit{Dept. of Computer Science and Engineering} & \textit{Univ Rennes, CNRS, Inria, IRISA} & \textit{Dept. of Computer Science and Engineering} \\
\textit{IIT Dharwad, India} & \textit{Rennes, France} & \textit{IIT Dharwad, India} \\

212011004@iitdh.ac.in & marcello.traiola@inria.fr & gayathri@iitdh.ac.in
\end{tabular}
}

\maketitle
\thispagestyle{firstpage}
\begin{abstract}
Deep neural network (DNN) inference at the edge demands simultaneous improvements in accuracy, computational efficiency, and energy consumption. Approximate computing and Mixture-of-Experts (MoE) architectures have each been studied as independent routes towards efficient inference, the former by replacing exact arithmetic with low-power approximate multipliers, the latter by routing inputs through specialized expert sub-networks to enable conditional computation. However, their interaction remains entirely unexplored. This paper presents AxMoE, the first study of the impact of approximate multiplication on MoE DNN architectures. We evaluate three MoE variants: \textit{Hard MoE, Soft MoE,} and \textit{Cluster MoE} against dense baselines across three CNN architectures \textit{(ResNet-20, VGG11\_bn, VGG19\_bn)} on CIFAR-100 and a Vision Transformer (ViT-Small) on Tiny ImageNet-200 dataset, using eight 8-bit signed multipliers (including one exact baseline) from the EvoApproxLib library. Results show that, without retraining, the Dense baseline is the most resilient topology across all CNN architectures, whereas on ViT-Small, all topologies degrade at comparable rates regardless of routing strategy. After approximate-aware retraining, recovery varies substantially across architectures, topologies, and multipliers. ResNet-20 achieves full recovery across the entire multiplier range, whereas VGG architectures recover at moderate multipliers but fail irreversibly at aggressive ones for all topologies except Cluster MoE on VGG11\_bn; on ViT-Small, Hard MoE outperforms Dense under aggressive approximation at equal normalized inference cost. These results pave the way for future approximate MoE hardware-software co-design strategies.

\end{abstract}


\section{Introduction}
\label{sec:intro}

From embedded vision to autonomous systems, deep neural network (DNN) inference demands immense computational resources, strictly limiting their deployment on energy-constrained edge devices. To mitigate this, hardware and software designers have historically relied on two complementary efficiency paradigms. At the hardware level, Approximate Computing (AxC) systematically trades numerical precision for substantial gains in energy efficiency and silicon area by employing reduced-accuracy arithmetic circuits (e.g., approximate multipliers)~\cite{jiang_approximate_2020}. At the DNN architecture level, the Mixture-of-Experts (MoE) approach introduces dynamic sparsity by routing inputs to specialized, localized sub-networks, decoupling the model's total parameter count from its active computational cost~\cite{jj-moe}.

Despite the maturity of both fields, they have been studied entirely in isolation. The approximate computing ecosystem has primarily targeted dense, statically-activated architectures in which every parameter and computational node is exercised by every input~\cite{rasheed_toward_2025}. MoE architectures fundamentally differ from this, as only a fraction of parameters are active per input, error propagation is gated by routing decisions, and the effective approximation fraction varies across the input distribution. To address this gap, this paper presents \textit{AxMoE}, the first comprehensive study of the impact of approximate multiplication on different dynamic MoE topologies.

Our study spans three distinct MoE paradigms: \textit{Hard MoE}~\cite{top1-moe,gshard,switch-moe}, \textit{Soft MoE}~\cite{soft-moe1}, and \textit{Cluster MoE}~\cite{cluster1,cluster2}, integrated into four widely used baseline architectures (ResNet-20, VGG11\_bn, VGG19\_bn, and ViT-Small). More specifically, the mentioned MoE paradigms represent two fundamentally different approaches to routing: Hard MoE 
and Soft MoE utilize a lightweight, per-layer router learned end-to-end, whereas Cluster MoE employs a standalone gateway network that routes entire inputs at the image level. By emulating eight Pareto-optimized 8-bit signed approximate multipliers from the EvoApproxLib library~\cite{evo}, we evaluate how these distinct routing topologies interact with arithmetic approximation. We first analyze the impact of the approximation on MoE without retraining or fine-tuning, and then apply approximation-aware retraining. The first analysis thereby reveals the intrinsic resilience to approximation error of each routing approach, a property that has not been previously characterized. The second analysis explores the trade-offs between accuracy and power consumption across the different approaches. Comparing pre- and post-retraining results reveals that the ability to recover lost accuracy is not uniform across models. Instead, the recovery gap (the amount of accuracy regained through retraining) varies significantly depending on the approximate multiplier, the routing strategy, and the underlying network architecture. 
    Without retraining, Dense architecture is the most error-resilient topology across all CNN architectures. On ViT-Small, for all topologies, Dense, Hard, Soft, and Cluster degrade at comparable rates, as MoE routing is confined to Feed Forward Network layers, while the Multi Head Self-attention linear projections, which account for approximately one-third of all effective MACs, are identically approximated in every variant.
    After approximate-aware retraining, ResNet-20 achieves full accuracy recovery across all eight multipliers for all topologies. On VGG architectures, full accuracy is recovered for most approximate multipliers, except for the most approximate ones (KVA and L2L). On ViT-Small, Hard MoE applied to 50\% of the layers exhibits the flattest degradation profile in the study, and Hard MoE applied to 25\% of the layers is the only configuration to exceed Dense accuracy at equivalent normalized inference cost under aggressive approximation.
    Finally, regarding the trade-off between accuracy and power consumption, we observe that the Dense model is always among the Pareto-optimal points across all explored NN architectures, whereas Soft MoE improves accuracy but incurs higher power consumption.
    
\section{State of the art and background}
\label{sec:sota}
\subsection{Approximate Computing for DNN Inference}
Deep Neural Networks (DNNs) are inherently error-resilient, making them prime candidates for approximate computing, a paradigm that trades numerical precision for significant gains in energy efficiency and hardware area. In DNN inference accelerators, the multiply-accumulate (MAC) operations dominate power consumption, motivating the deployment of approximate multipliers that utilize reduced-accuracy arithmetic logic circuits.  The EvoApproxLib library~\cite{evo}  provides a comprehensive benchmark set of Pareto-optimized approximate circuits spanning diverse power-accuracy trade-offs, and serves as the basis for emulating approximate hardware in this work.

To bridge the gap between custom hardware designs and high-level deep learning software, several emulation and training frameworks have been developed. Early work such as  AxNN~\cite{axnn} demonstrated backpropagation based identification of resilient neurons for selective approximation. ALWANN~\cite{alwann} introduced automated layer-wise approximate multiplier assignment for CNNs without retraining. TFapprox~\cite{tfapprox} accelerated inference simulation using lookup tables (LUTs), while ApproxTrain~\cite{approxtrain} extended fast GPU-accelerated LUT simulations to full DNN training cycles. AdaPT~\cite{adapt} introduced a robust fine-tuning framework that incorporates approximate arithmetic directly into the retraining loop. TransAxx~\cite{transaxx} extended these capabilities to transformer architectures, enabling approximate-aware fine-tuning for Vision Transformers. All prior work targets dense, monolithic DNN architectures, and, to the best of our knowledge, none has examined the intersection of approximate arithmetic and dynamic MoE routing, which is the central contribution of AxMoE.

\subsection{Mixture of Experts Architectures}
\label{subsec:moe}
MoE augments DNNs by partitioning computation across multiple specialized sub-networks, or experts, to improve model capacity while maintaining low per-input computational cost~\cite{jj-moe}. A router (or gating network) dynamically determines which expert(s) to invoke for each input. Formally, for a given input~\texttt{x}, the output~\texttt{y} is computed as a weighted combination of expert outputs and is given by 
 $y(x)= \sum_{i}^{n} G_{i}(x)E_{i}(x)$, where $E_{i}(x)$ is the output of the $i^{th}$ expert and $G(x)=Softmax(Wx)$ is the router. $W$ in $G(x)$ is the trainable weight matrix of the Router.
 
This work investigates three MoE variants. In the \texttt{Hard MoE} design variant~\cite{top1-moe,gshard,switch-moe}, the router selects the single highest-scoring expert~\texttt{(Top-1 routing)}, i.e., $y(x)= G_{i}(x)E_{i}(x)$ where $i$ is the index of the expert maximizing $G(x)$. 
This sparse activation strategy underlies state-of-the-art MoE models like Sparsely-Gated Mixture-of-Experts~\cite{top1-moe}, Gshard~\cite{gshard}, Switch transformer~\cite{switch-moe}. In  \texttt{Soft MoE}~\cite{soft-moe1}, every expert contributes to each output, with $y(x)= \sum_{i}^{n} G_{i}(x)E_{i}(x)$, producing more stable training at the cost of full expert activation per sample.  Both \texttt{Hard} and \texttt{Soft} MoE variants embed a lightweight router at each MoE layer, a single linear projection followed by a softmax, learned jointly with the rest of the network, and adding negligible parameter overhead per layer. In \texttt{Cluster MoE}~\cite{cluster1,cluster2}, the routing philosophy is fundamentally different. Each expert is a full replica of the base dense model, fine-tuned independently on a disjoint subset of classes. Rather than a per-layer gate, routing is handled by a dedicated standalone gateway network that processes the full input image and selects the appropriate expert before inference begins. Unlike per-layer routers, this one-time routing decision incurs a fixed overhead regardless of model depth, but the gateway must be expressive enough to reliably partition the input space, making the gateway complex to design.

Because MoE models route inputs to localized sub-networks based on data-dependent criteria, the error propagation mechanisms under approximate arithmetic fundamentally diverge from those of dense networks. In Hard MoE, only the selected expert's computation is subjected to approximation on any given input. In Soft MoE, all experts execute for every input, so the approximated outputs of all experts contribute to the final result. 
No prior study has characterized how these structural differences shape power–accuracy trade-offs under approximate hardware, motivating the \textit{AxMoE} framework.

\section{Proposed Methodology}
\label{sec:method}

\subsection{MoE Layer Approximation}
\label{subsec:substitution}

To evaluate the AxMoE framework across diverse topologies, we perform
targeted MoE layer substitution on each dense baseline, preserving dimensional
compatibility while avoiding macroscopic architectural changes.
The substitution strategy is tailored to each architecture's topology. In ResNet-20, we target the convolutional pathways within the residual blocks: a lightweight router selects the appropriate convolutional expert for each
input feature map, while the residual skip connections remain untouched. In VGG11\_bn and VGG19\_bn, which lack skip connections, we directly substitute the deep \texttt{Conv2d} layers with MoE layers. In ViT-Small, MoE conversion is conventionally confined~\cite{gshard,switch-moe} to the Feed-Forward Networks (FFN) within a specified fraction of transformer blocks (ratio $\in$ $\{0.25, 0.5\}$), as they contain the vast majority of the model's parameters and process tokens independently. This isolates the conditional computation from the Multi-Head Self-Attention (MSA) layers, where the inherent need for global token mixing makes sparse routing computationally prohibitive. The Multi-Head Self-Attention (MSA) and patch-embedding modules execute densely for every patch, while the router dynamically distributes patches among the $N$ FFN experts within each converted block.

The layer type subjected to approximate multiplication varies across architecture families, and this choice is motivated by architectural considerations. For CNN architectures (ResNet-20, VGG11\_bn,  VGG19\_bn), approximate multiplication is applied to Conv2d layers, which account for 98–99.9\% of effective MACs in these models. Linear (fully-connected) layers and BatchNorm operations remain exact.
For ViT-Small, the transformer body contains no Conv2d operations beyond the single patch-embedding layer (which contributes less than 1\% of total MACs and is kept exact). The dominant computations are entirely in nn.Linear layers: specifically, the QKV projection, multi-head output projection, and both FFN layers (fc1, fc2) within each transformer block. These collectively account for approximately 97.3\% of the effective MACs. Approximate multiplication for ViT architecture,  therefore targets all linear layers within the transformer blocks, excluding the classifier head.
In all variants, the routing gate executes with exact arithmetic; only the expert computation is subjected to approximate multiplication.

\subsection{Computational Cost Metrics}
\label{subsec:metrics}

\noindent\textbf{Static vs. Effective MACs.}
\textit{Total (Static) MACs} represent the worst-case computational capacity of the model, the sum of MACs across all experts plus any gateway network.
\textit{Effective MACs} ($M_{\mathrm{eff}}$) represent the dynamic cost of a single inference pass: for MoE architectures, this is the sum of the \emph{selected} expert's MACs and the router's MACs. In standard dense networks, Static MACs and Effective MACs are identical (e.g., $41.63$\,M for ResNet-20), since the entire network executes for every input.

\smallskip
\noindent\textbf{MAC formulation.}
For CNNs, the total MAC count aggregates operations across all convolutional and
fully-connected (linear) layers:
\begin{equation}
  M_{\mathrm{total}} = \sum_l M_{\mathrm{conv}}(l) + \sum_l M_{\mathrm{linear}}(l),
\end{equation}
where the convolutional MACs for a single layer are
$C_{\mathrm{out}} \times H_{\mathrm{out}} \times W_{\mathrm{out}}
 \times (C_{\mathrm{in}}/G) \times k_H \times k_W$.
Since only convolutional layers are subjected to approximate multiplication
(linear and batch-normalization layers remain exact), the approximate MAC
fraction is
\begin{equation}~\label{eq:fapx}
  f_{\mathrm{apx}} = \min\!\left(\frac{M_{\mathrm{approx}}}{M_{\mathrm{eff}}},\; 1.0\right),
\end{equation}
where $M_{\mathrm{approx}}$ equals $M_{\mathrm{conv}}$ for Dense and
Hard MoE, and $n_{\mathrm{exp}} \times M_{\mathrm{conv}}$ for Soft MoE. 

\smallskip
\noindent\textbf{Effective MACs per variant.}
For Dense networks, every layer executes for each input, so
$M_{\mathrm{eff}}^{\mathrm{dense}} = M_{\mathrm{total}}$.
For Hard MoE, a lightweight per-layer gate routes each input to exactly one
expert, so the effective cost is a single backbone pass plus negligible routing
overhead:
$M_{\mathrm{eff}}^{\mathrm{hard}} = M_{\mathrm{backbone}} + M_{\mathrm{gate}}$.
For Soft MoE, all $n_{\mathrm{exp}}$ experts process every input and their
outputs are blended, giving
$M_{\mathrm{eff}}^{\mathrm{soft}} = n_{\mathrm{exp}} \times M_{\mathrm{total}}^{\mathrm{backbone}}$,
which incurs a ${\sim}3\times$ increase in Effective MACs and GPU memory
footprint relative to Dense.
Cluster MoE routes at the image level via a standalone gateway network
(${\sim}125.8$\,M MACs for CNNs; ${\sim}4.14$\,G MACs for ViT-Small) that is
shared across experts; its cost is added once per inference pass.

\subsection{Power Normalisation}
\label{subsec:power}

To enable unified cross-architecture comparison, we express total inference
power as a ratio normalized to the reference architecture, i.e., dense (i.e., non-MoE) model executed with an 8-bit signed precise multiplier (mul8s\_1KV6 from EvoApproxLib). In detail, let $M_{\mathrm{base}}$ denote the Effective MACs of the reference architecture,
$f_{\mathrm{apx}}$ the approximate MAC fraction defined in Eq.~\eqref{eq:fapx}, and
$P_{\mathrm{apx}}/P_{\mathrm{KV6}}$ the relative power of the chosen multiplier
with respect to the exact KV6 baseline.
The normalized power is then
\begin{equation}
  P_{\mathrm{norm}}
  = \frac{M_{\mathrm{eff}}}{M_{\mathrm{base}}}
    \!\left[\,
      f_{\mathrm{apx}}\,\frac{P_{\mathrm{apx}}}{P_{\mathrm{KV6}}}
      + \left(1 - f_{\mathrm{apx}}\right)
    \right].
  \label{eq:pnorm}
\end{equation}
 $P_{\mathrm{norm}} = 1.0$ for the reference baseline Dense(KV6).
$P_{\mathrm{norm}} < 1.0$ indicates a net power saving relative to that
reference; $P_{\mathrm{norm}} > 1.0$ indicates additional overhead.
The term $f_{\mathrm{apx}}\,\frac{P_{\mathrm{apx}}}{P_{\mathrm{KV6}}}
      + \left(1 - f_{\mathrm{apx}}\right)$ in~\eqref{eq:pnorm} is the power-normalized cost per
effective MAC: it blends the approximate multiplier's reduced power (weighted by
the approximated fraction) with the exact power of the remaining operations.
The leading ratio $M_{\mathrm{eff}}/M_{\mathrm{base}}$ then captures any
MAC-count overhead introduced by the routing topology.

\subsection{Emulation Framework and Retraining }
\label{subsec:emulation}

Approximate 8-bit signed multiplication is emulated via pre-computed
look-up tables (LUTs) that map every pair of 8-bit operands to the output of
the target approximate multiplier, enabling efficient GPU-parallel inference
without custom hardware.
For CNN MoE variants, we use a GPU-accelerated LUT-based inference engine
following the approach of TFApprox~\cite{tfapprox}.
For ViT-Small, we use the TransAxx framework~\cite{transaxx}, which extends
LUT-based approximate emulation to transformer architectures and integrates
approximate-aware fine-tuning within the PyTorch training loop.

\section{Experimental Evaluation}

\subsection{Experimental Setup}
We conduct all our experimental evaluations on NVIDIA GPUs (RTX A6000 and RTX 4000 Ada). Table~\ref{tab:dnn models} provides the details and characteristics of the DNN models used in this work. Table~\ref{tab:axxmults} provides the details of the approximate multipliers used in this work. We obtain the baseline pretrained models for the CNNs from~\cite{cifar_models} and use \texttt{vit\_small\_patch16\_224} from Pytorch \texttt{timm} library with finetuning as the baseline transformer model~\cite{vit_models}.
\begin{table}[!htb]
    \caption{DNN Models along with Characteristics.}
    \label{tab:dnn models}
    \centering
    \begin{tabular}{c|P{1.8cm}|c|c}
    \textbf{Model Name} &  \makecell{\textbf{Model Layer}\\\textbf{Architecture}} & \makecell{\textbf{No. of}\\\textbf{Parameters}\\\textbf{ (M)}} & \textbf{Dataset}\\
    \hline
      ResNet-20   & 21 Conv2D and 1 Linear & 0.28 & CIFAR-100 \\
    \hline
      VGG11\_bn    & 8 Conv2D and 1 Linear & 9.803 & CIFAR-100 \\
    \hline
      VGG19\_bn   & 16 Conv2D and 1 Linear & 20.612 & CIFAR-100\\
    \hline
      ViT-Small  & 12 Transformer Blocks  & 21.743 & Tiny ImageNet\\
    \end{tabular}
\end{table}

\begin{table}[!htb]
     \caption{Details of Approximate Multipliers from EvoApproxLib~\cite{evo}. }
     \label{tab:axxmults}
     \centering
     \begin{tabular}{c|c|c|c}

    \textbf{Multiplier} & \makecell{\textbf{Power}\\\textbf{(nW)}} & \makecell{\textbf{Per-Op}\\\textbf{Saving (\%)}} & \makecell{\textbf{Error}\\\textbf{Probability (\%)}} \\
    \hline
      mul8s\_1KV6 (exact)   &  0.425 & 0.0 & 0.0 \\
    \hline
     mul8s\_1KV8     &  0.422 & 0.7 & 50 \\
    \hline
     mul8s\_1KV9     & 0.410 & 3.5  & 68.75 \\
    \hline
    mul8s\_1KVA    &  0.391 & 8.0  &81.25  \\
    \hline
     mul8s\_1KVM     &  0.369 & 13.2 & 49.80 \\
    \hline
     mul8s\_1KVP     &  0.363 & 14.6  & 74.8 \\
    \hline
     mul8s\_1L2J     &0.301  & 29.2 & 74.61 \\
    \hline
      mul8s\_1L2L     & 0.200 & 52.9  & 93.16\\
     \end{tabular}
     \vspace{-0.5cm}
 \end{table}
 
\textit{Datasets and Training Parameters:} For CNNs, experiments were conducted on the CIFAR-100 dataset~\cite{cifar_data}, which consists of 50,000 training and 10,000 test RGB images of size $32\times32$, grouped into 100 classes. For ViTs, experiments were conducted on the Tiny ImageNet dataset~\cite{tinyimage_data}, which consists of 100,000 training images, 10,000 validation images, and 10,000 test images of size $64\times 64$, grouped into 200 classes.
For training of the MoE variants, we use \textit{SGD optimizer} with a learning rate of $0.1$ and an L2 regularization weight decay rate of $5e^{-4}$ and train for \textit{200 epochs} with a \textit{batch size of 128}.

\textit{Approximate-aware retraining}: We apply
approximate-aware retraining for 5 epochs using SGD (learning rate $0.1$,
L2 weight decay $5e^{-4}$, batch size $128$).
In all configurations, the routing gate parameters are frozen during retraining
so that only the expert weights are updated, keeping the routing decisions
independent of the approximate hardware. 

We retrain for 5 epochs, as LUT-based approximate-aware retraining resolves every multiplication through a 256×256 table, incurring $3–5\times $ the wall-clock cost of standard fine-tuning~\cite{transaxx}. Each CNN epoch requires 6–8 minutes per architecture-variant pair on an RTX 4000 Ada ($\approx 30-40$ minutes per 5-epoch run; $\approx48–64$ GPU-hours for the full 96-configuration CNN sweep); each ViT epoch requires 45–50 minutes. Despite the short epoch budget, full accuracy recovery is achieved for all ResNet-20 variants and for all VGG variants at multipliers KVM–L2J (Fig.~\ref{fig:all_archs_apx_acc}), confirming that 5 epochs is sufficient for the regime where recovery is possible, and that the catastrophic failures at KVA/L2L were not observed to resolve with additional epochs, suggesting a limitation that is independent of training budget.

\section{Results}

\subsection{Baseline Results with Exact Multiplier}
\label{sec:res-exact}
Table~\ref{tab:baseline-results-exact} reports Top-1 accuracy, active parameters, total and effective MACs, and GPU memory for all variants under exact multiplication (mul8s\_1KV6). The effective MAC counts are based on the MAC cost model described in Section~\ref{subsec:metrics}. Hard MoE effective MACs are marginally higher than Dense, with router overhead adding less than 0.02\% (e.g., for ViT-Small, 4245.35 M vs. 4244.66 M), since Top-1 routing executes a single expert per input. Soft MoE incurs $2.96–2.99\times$ overhead on CNNs and $1.33–1.66\times$ on ViT-Small as all experts process every input. Cluster MoE overhead ranges from $1.31\times$ (VGG19\_bn) to $3.96\times$ (ResNet-20), driven by its standalone gateway network. These overheads directly set the normalized power baseline before any approximate multiplier is applied. Soft and Cluster MoE must therefore achieve disproportionately large per-operation savings when approximated to match the efficiency of their Dense and Hard MoE counterparts.

Under exact arithmetic, Soft MoE delivers the highest CNN Top-1: a $2.3$ percentage points (pp) increase in accuracy over Dense on ResNet-20 (70.88\% vs. 68.58\%) and 0.66\pp  on VGG11\_bn, confirming that all-expert routing benefits shallow networks. The advantage reverses for deeper models. VGG19\_bn Soft MoE is less accurate than Dense by 1.62\pp (71.69\% vs. 73.31\%), indicating that all-expert routing introduces optimization difficulty when the network is already over-parameterized relative to its task. Hard MoE incurs an accuracy penalty that grows with network depth. We observe the following reductions in accuracy: 3.1\pp (ResNet-20), 2.79\pp (VGG11\_bn) and 5.41\pp (VGG19\_bn). On ViT-Small, Hard MoE r=0.25 trails Dense by only 0.29 pp (81.17\% vs. 81.46\%), and Soft r=0.25 marginally exceeds Dense at 81.50\%, indicating that MoE routing within FFN layers has a negligible impact on ViT-Small accuracy under exact arithmetic. For ViT-Small, Cluster MoE suffers the largest baseline penalty (72.38\%), a 9.08\pp reduction at 1.98$\times$ effective MACs.
\begin{table}[!htb]
    \centering
    \caption{Baseline Results with Exact Multiplier. }
    \label{tab:baseline-results-exact}
    \resizebox{\columnwidth}{!}{
    \begin{tabular}{l|c|c|c|c|c|r}
      \makecell{\textbf{DNN}\\\textbf{Model}}  & \makecell{\textbf{MoE}\\\textbf{Type}} & \makecell{\textbf{Active}\\\textbf{Params}\\\textbf{(M)}} & \makecell{\textbf{Total}\\\textbf{MACs}\\\textbf{(M)}} & \makecell{\textbf{Eff.}\\\textbf{MACs}\\\textbf{(M)}} & \makecell{\textbf{Top-1}\\\textbf{(\%)}} & \makecell{\textbf{GPU}\\\textbf{Mem}\\\textbf{(MB)}}  \\
      \hline
      \hline
      \multirow{5}{*}{\textbf{ResNet-20}} & \textbf{\texttt{Dense}} & 0.27 & 41.63 & 41.63 & 68.58  & 43.8 \\
                                & \textbf{\texttt{Hard}} & 0.28 & 123.25  & 41.63 & 65.48 & 62 \\
                                & \textbf{\texttt{Soft}} &  1.091 & 123.25 & 123.25 & 70.88 & 119 \\
                                & \textbf{\texttt{Cluster}} & 1.13  & 250.69 & 164.73 & 68.33 & 43.8 \\
                               
      \hline
      \multirow{5}{*}{\textbf{VGG11\_bn}} & \textbf{\texttt{Dense}} & 9.80 & 153.95 & 153.95 & 70.35  & 112.7 \\
                                & \textbf{\texttt{Hard}} & 9.81 &   458.96 & 153.95 & 67.56 & 182.8 \\
                                & \textbf{\texttt{Soft}} &  36.934 & 458.96 & 458.96 & 71.01 & 472.2 \\
                                & \textbf{\texttt{Cluster}} &  10.66 & 587.53 & 279.77 & 68.9 & 112.7 \\
      \hline
      \multirow{5}{*}{\textbf{VGG19\_bn}} & \textbf{\texttt{Dense}} & 20.61 & 399.92 & 399.92 & 73.31  & 186.1 \\
                                & \textbf{\texttt{Hard}} & 20.62 &   1195.67 & 399.92 & 67.9 & 401.7 \\
                                & \textbf{\texttt{Soft}} &  80.153 & 1195.67 & 1195.67 & 71.69 & 669.1 \\
                                & \textbf{\texttt{Cluster}} &  21.462 & 1325.45 & 525.69 & 68.83 & 186.1 \\
      \hline
      \multirow{7}{*}{\textbf{ViT-Small}} & \textbf{\texttt{Dense}} & 21.743 & 4244.66 & 4244.66 & 81.46  & 609.3 \\
                                & \textbf{\texttt{Hard}} (0.25) & 21.746 &   5641.51 & 4245.35 & 81.17 & 637.6 \\
                                & \textbf{\texttt{Soft}} (0.25) &  28.836 & 5641.51 & 5641.51 & 81.5 & 712 \\
                                & \textbf{\texttt{Hard}} (0.5) & 21.75 &   7038.36 & 4246.04 & 77.49 & 665.8 \\
                                & \textbf{\texttt{Soft}} (0.5) &  28.836 & 7038.36 & 7038.36 & 78.17 & 740.3 \\
                                & \textbf{\texttt{Cluster}} &  36.713 & 16873.8 & 8384.66 & 72.38 & 711.3 \\
    \hline
    \hline
     
    \end{tabular}
    }
    \vspace{-0.25cm}
\end{table}

\subsection{Results with Approximate Multipliers}
\label{sec:res-approx}

\begin{figure*}[!t]
      \centering
	   \includegraphics[width=0.9\linewidth]{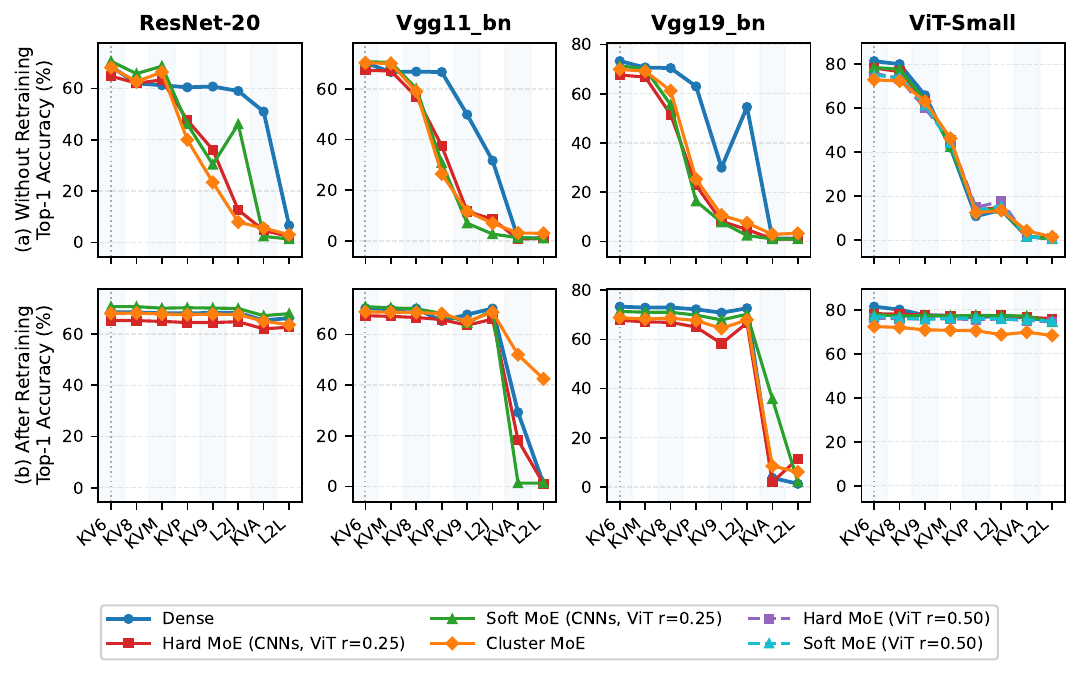}
	\caption{Accuracy Degradation vs Approximation Aggressiveness. Part (a) of the figure provides the Absolute Top-1 Accuracy without retraining, and (b) provides Absolute Top-1 Accuracy with retraining for 5 epochs.}
	\label{fig:all_archs_apx_acc}
\end{figure*}
\begin{figure*}[!h] 
\centering
\includegraphics[width=0.9\linewidth]{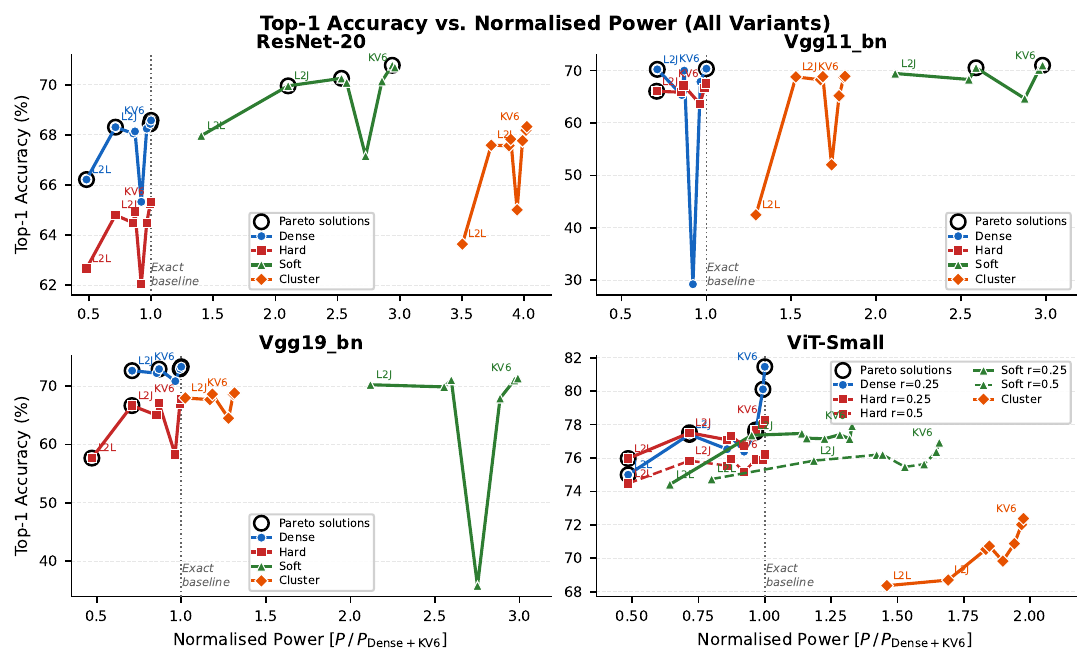}
    \caption{Top-1 Accuracy (after approximate-aware retraining) vs Normalised Power Consumption w.r.t Dense Exact Multiplier across different MoE Variants.}
    \label{fig:top-1_vs_pow}
    \vspace{-0.45cm}
\end{figure*}

Without retraining, the exact-arithmetic accuracy hierarchy partially inverts (Fig.~\ref{fig:all_archs_apx_acc}(a)). 
\textbf{For CNN architectures}, Dense baseline is the most intrinsically resilient topology across all CNN architectures: on ResNet-20, Dense retains 59.1\% at L2J while Hard MoE falls to 12.6\% and Cluster MoE to 9.4\%. All VGG variants collapse below 20\% accuracy by KV9 (3.5\% per-operation saving) except Dense, which sustains accuracy through L2J. ResNet-20 is categorically more robust than VGG across every topology, with all four variants remaining functional through L2J; no VGG variant except Dense retains acceptable accuracy beyond KV8. A localized anomaly appears at KVM (13.2\% saving), where Soft MoE on all three CNN architectures maintains near-KV6 accuracy (within 0.1–1.9 pp of exact baseline). 

On \textbf{ViT-Small}, all variants degrade at almost the same rate. 
This is because in ViT-Small, approximate multiplication is applied to all linear projection layers inside each transformer block. This includes both the FFN layers (which are controlled by MoE routing) and the QKV and output projection layers within the self-attention module (which are always Dense, regardless of MoE configuration). The self-attention projections alone account for roughly one-third of all multiply-accumulate operations. Because these layers are approximated identically in every variant, Dense, Hard, Soft, and Cluster all have the same self-attention structure, and every variant is exposed to the same source of error before the FFN is even reached. The net result is that the routing strategy has negligible influence on pre-retraining accuracy for ViT-Small.

After approximate-aware retraining (Fig.~\ref{fig:all_archs_apx_acc}(b)), results change substantially. ResNet-20 achieves full recovery across the entire multiplier range for all variants, including KVA and L2L. 
For example, when the L2J multiplier is used, the Dense variant achieves a 0.27 pp (68.31\%) reduction, whereas Soft MoE achieves a 0.92 pp (69.96\%) reduction. The residual skip connections that provide pre-retraining resilience also stabilize approximate-aware learning. 
On VGG architectures, the accuracy is fully restored for most multipliers in Dense, Hard, and Soft; only with KVA and L2L multipliers is it not recovered through retraining for any of these three topologies. In VGG11\_bn, Cluster MoE is the sole exception, retaining 52.0\% with KVA and 42.5\% at L2L.
VGG19\_bn Hard MoE exhibits a non-monotonic response to approximation aggressiveness, the accuracy collapses to 2\% at KVA (81.25\% error probability) yet partially recovers to 57.66\% at the more aggressive L2L (93.16\% error probability).
This indicates that error probability alone does not determine retraining outcome, and that the two multipliers present qualitatively different retraining conditions.

On ViT-Small, after retraining, Dense degrades more steeply than the others when more approximate multipliers are applied, losing 6.46 pp with L2L w.r.t. KV6, while Hard r=0.5 loses only 1.74 pp across the full multiplier range and has the flattest degradation profile. 
Unlike CNN Hard MoE, ViT routes at the patch level and each patch is independently routed per block, so each expert adapts its weights to a narrow, consistent subset of patch representations during retraining. Hard MoE for CNNs routes at the image level, so each expert still processes full spatially diverse feature maps and gains no equivalent adaptation advantage. Soft MoE degrades more gradually than Dense when measured from each variant's own exact baseline: Soft MoE r=0.25 loses 3.50 pp to L2L (77.92\% to 74.42\%) versus Dense's 6.46 pp drop (81.46\% to 75.00\%).

\subsection{Accuracy vs Power Consumption}
\label{sec:res-top1-acc}

Fig.~\ref{fig:top-1_vs_pow} plots post-retraining Top-1 accuracy against normalized power for all variants. The points highlighted with black circles on each panel form the Pareto-optimal frontier: the set of operating points from which no other point simultaneously achieves lower power and higher accuracy. It traces the best achieved accuracy across all power levels. MoE topology overhead is the dominant factor shaping the power envelope; while approximate multipliers help reduce it, they usually penalize accuracy. On ResNet-20 and both VGG architectures, Dense and Hard MoE both reduce power relative to the baseline (vertical line), with approximate multipliers reducing normalized power to $0.71\times$ at L2J and $0.47–0.48\times$ at L2L. Dense+L2J is Pareto-optimal for all three CNN architectures: 68.31\% at 0.71$\times$ (ResNet-20), 70.22\% at 0.71$\times$  (VGG11\_bn), 72.64\% at 0.71$\times$  (VGG19\_bn), each with $\leq$ 0.70 pp accuracy loss. Soft MoE improves accuracy for ResNet-20 and VGG11\_bn but incurs high power consumption; indeed, when using the lowest-power multiplier (L2L), it still exceeds the Dense baseline with KV6. 
On ViT-Small, Hard MoE r=0.25 and r=0.5 both have $\approx1.0\times$ baseline power (effective MACs $\approx$ Dense). With L2L, Hard MoE r=0.25 achieves 75.97\% accuracy versus Dense's 75.00\% at equivalent normalized power. 
Soft MoE variants ($1.33\times$ power) offer no Pareto advantage over Hard at any multiplier.

The Cluster MoE variant is always entirely Pareto-dominated across all experiments. For example, for ViT-Small, at its most aggressive operating point (L2L, $1.46\times$ normalized power), it achieves 
only $68.36\%$, while Hard~$r{=}0.25$ reaches $75.97\%$ at $1.0\times$ power, $7.61$\pp higher accuracy at half the cost.

\subsection{Pareto Optimal Analysis}

\textbf{CNN architectures.}
On ResNet-20, the Dense baseline occupies four of the seven Pareto-optimal points (L2L, L2J, KV8, KV6), confirming it as the variant of choice across the entire power-saving range. Soft MoE occupies three 
Pareto-optimal points at $2.1\times$--$2.9\times$ power (KV8, KVP, 
L2J), providing marginally higher peak accuracy ($70.78\%$ vs. $68.58\%$) but only at power levels that exceed the Dense+KV6 exact 
baseline by $2\times$. On ResNet-20, Hard MoE and Cluster MoE are entirely Pareto-dominated.

On VGG11\_bn, the frontier comprises five points. Hard+L2J ($0.71\times$, $66.04\%$), Dense+L2J ($0.71\times$, 
$70.22\%$), and Dense+KV6 ($1.0\times$, $70.35\%$). Soft MoE contributing two further points at $2.6–3.0\times$ power (KVM: 70.52\%; KV6: 70.99\%), offering marginal accuracy gains over Dense+KV6 at a higher power cost.
 
On VGG19\_bn, the frontier extends to six points, with Hard MoE contributing two (L2L at $0.47\times$, L2J at $0.71\times$) in the aggressive savings region.
Across all three CNN architectures, the Dense variant using the L2J multiplier is always on the Pareto-front, delivering 29.2\% per-operation savings with less than $0.70$\,pp accuracy loss.

\textbf{ViT-Small.}
The ViT Pareto frontier spans eight points, though six are visually distinct in Fig.~\ref{fig:top-1_vs_pow}: at L2L, L2J, and KV9, Dense r=0.25 and Hard MoE r=0.25 differ by less than 0.001× in normalized power and appear as coincident markers. It exhibits a qualitatively different topology (Fig.~\ref{fig:top-1_vs_pow}, bottom right).  Dense r=0.25 and Hard MoE r=0.25 alternate on the frontier at every multiplier from L2L to KV6: at L2L ($0.485\times$ power), Hard MoE r=0.25 achieves $75.97\%$ versus Dense's $75.00\%$, making it the \emph{only} configuration in this study where an MoE topology achieves strictly higher accuracy than Dense at strictly lower power. The advantage persists at L2J and KV9, where Hard MoE r=0.25 leads Dense by $0.10$--$0.13$\,pp at effectively identical normalized power. Neither Soft MoE nor Cluster MoE appears on the ViT Pareto frontier: Soft (r=0.25, $1.33\times$ power) is dominated by Hard MoE r=0.25 at every multiplier, and Cluster ($1.46\times$ power) is dominated by Hard MoE r=0.25 at half the normalized power with $7.61$ \pp higher accuracy at L2L.

\section{Conclusion}
This paper presents AxMoE, the first empirical study of the impact of approximate multipliers on MoE DNN architectures. Across four architectures, three MoE topologies, and eight approximate multipliers, results show that the interaction between routing strategy and arithmetic approximation is strongly architecture-dependent and cannot be inferred from prior dense-network results alone.

Without retraining, the Dense model is the most resilient topology across all CNN architectures, retaining acceptable accuracy even with moderately approximate multipliers, whereas MoEs degrade at significantly lower approximation. On ViT-Small, all topologies degrade at comparable rates prior to retraining. After approximate-aware retraining, ResNet-20 achieves full recovery across all multipliers. VGG architectures recover fully only when using moderate multipliers. 
Across all explored CNN architectures, the Dense variant with the L2J multiplier is always on the Pareto-front between accuracy and power consumption, delivering 29.2\% per-operation savings with an accuracy loss of under 0.70 pp. On ViT-Small, Hard MoE topologies exhibit substantially flatter post-retraining degradation profiles than Dense under aggressive approximation. We hope this study will pave the way and encourage future research on approximate MoE hardware-software co-design strategies.

\section*{Acknowledgments}
Part of this work was supported by Inria through the AxTRADE associate team and by the French National Research Agency (ANR) through the RADYAL project ANR-23-IAS3-0002 and REAxION project ANR-25-CE25-5926.

\bibliography{ref.bib}
\bibliographystyle{IEEEtran}

\end{document}